\documentclass{article}
\usepackage{multirow}
\usepackage{booktabs}
\usepackage{diagbox}
\usepackage{spconf,amsmath,epsfig}
\usepackage[colorlinks=true, allcolors=blue, urlcolor=purple]{hyperref}

\name{Adrien Chan-Hon-Tong\textsuperscript{1}, Gaston Lenczner\textsuperscript{1,2}, Aurelien Plyer\textsuperscript{1}\thanks{Thanks to the BPI AI4GEO project \textbf{http://ai4geo.eu/} for funding. \newline Corresponding author: \href{mailto:gaston.lenczner@alteia.com}{gaston.lenczner@alteia.com}}} 
\address{\textsuperscript{1}ONERA/DTIS, Universit{é} Paris-Saclay, FR-91123 Palaiseau, France\\\textsuperscript{2}Alteia, FR-31400 Toulouse, France} 
\usepackage{amssymb,amsmath}
\usepackage{graphicx}
\usepackage{epsfig}
\usepackage{url}
\usepackage{algorithm}
\usepackage{algorithmic}
\usepackage{tikz}
\usepackage[symbol]{footmisc}
\begin{document}
\title{Demotivate adversarial defense in remote sensing}

\author{}

\maketitle

\begin{abstract}


Convolutional neural networks are currently the state-of-the-art algorithms for many remote sensing applications such as semantic segmentation or object detection. However, these algorithms are extremely sensitive to over-fitting, domain change and adversarial examples specifically designed to fool them. 
While adversarial attacks are not a threat in most remote sensing applications, one could wonder if strengthening networks to adversarial attacks could also increase their resilience to over-fitting and their ability to deal with the inherent variety of worldwide data. In this work, we study both adversarial retraining and adversarial regularization as adversarial defenses to this purpose. However, we show through several experiments on public remote sensing datasets that adversarial robustness seems uncorrelated to geographic and over-fitting robustness. 
\end{abstract}

\section{Introduction}
World seen from remote sensing sensors can exhibit a great variability in appearance and the algorithms should be robust to these domain changes.
For example, a roof segmentation module~\cite{chen2018aerial} should deal with different roof appearances regardless of the country, weather, density area or time of the day. Currently, a deep learning module trained on some part of the world could perform very well on neighborhood regions while having poor performance far away~\cite{castillo2019data}.
This is an issue for both high resolution datasets which cover limited areas due to the cost of data and for low resolution datasets which either deal with high level label only or cover limited areas due to the ground-truth annotation cost. Deep neural networks are also notoriously prone to over-fit on training data. In such scenarios, data augmentation and regularization are common efficient ways to strengthen learning algorithms.

\begin{figure}[t]
   \begin{minipage}[t]{.24\linewidth}
   \centering\epsfig{figure=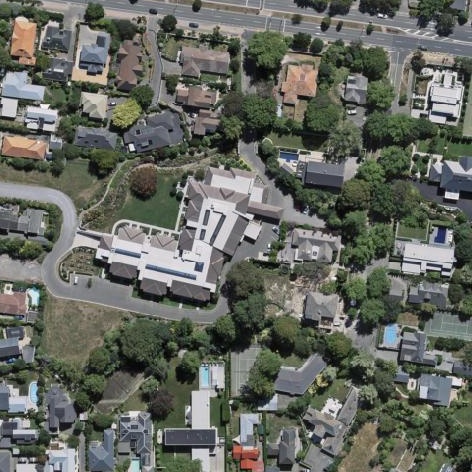, width=\linewidth}
    Original input
  \end{minipage} \hfill
   \begin{minipage}[t]{.24\linewidth}
   \centering\epsfig{figure=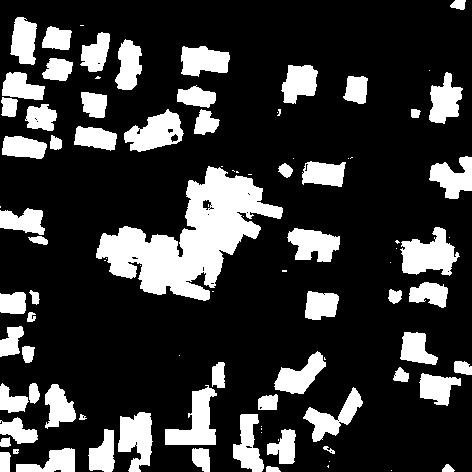,width=\linewidth}
    Original prediction
  \end{minipage} \hfill
   \begin{minipage}[t]{.24\linewidth}
   \centering\epsfig{figure=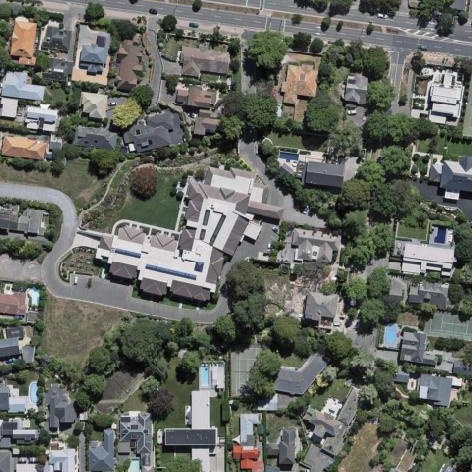,width=\linewidth}
    Attacked input
  \end{minipage} \hfill
    \begin{minipage}[t]{.24\linewidth}
   \centering\epsfig{figure=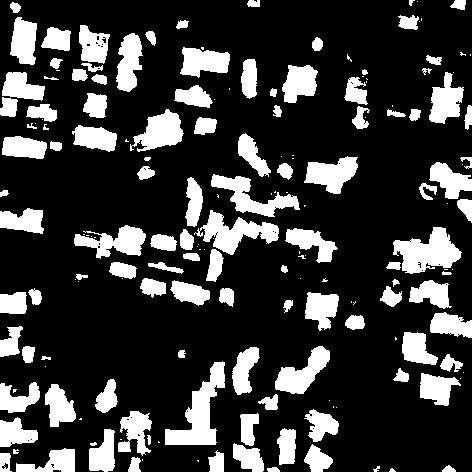,width=\linewidth}
    Attacked prediction
  \end{minipage} \hfill
  \caption{Adversarial attack on a network trained on AIRS~\cite{chen2018aerial} for building segmentation without adversarial defense}
  \label{fig:adv}
 \end{figure}

On the other hand, deep neural networks are also extremely weak against adversarial attacks~\cite{goodfellow2014explaining} which aim to find optimal examples to fool the algorithms. Figure~\ref{fig:adv} shows an example of the impact such attack. Adversarial defense has been widely studied in the literature to protect the networks from these attacks and these approaches usually rely on adversarial retraining~\cite{shaham2018understanding} or adversarial regularization~\cite{bartlett2017spectrally}. However, contrary to autonomous driving, adversarial attacks are not a real threat in most remote sensing applications which are neither in real time nor safety critical (e.g. land cover mapping) and due to the large physical changes implied. In this context, adversarial defense could be seen as being hard data augmentation and regularization and thus still be relevant by benefiting to geographic and over-fitting robustness.
\subsection{Contributions}
In this paper, we focus on evaluating whether adversarial defense is relevant to strengthen neural networks for remote sensing applications without adversarial examples threats. As a result, we found that adversarial robustness seems uncorrelated to geographical and over-fitting ones. 
Precisely, the contributions of this paper are: 
\begin{itemize}
    \item An evaluation of the impact of adversarial defense on robustness to geographic transfer, adversarial attacks and over-fitting.
    \item We practically study both adversarial retraining and adversarial regularization on semantic segmentation and detection tasks on public remote sensing datasets. 
    \item We show that adversarial robustness seems uncorrelated to transfer and over-fitting robustness.
\end{itemize}

Code is available\footnote{\href{https://github.com/achanhon/Lispchitz_penalty}{https://github.com/achanhon/Lispchitz\_penalty}} and contains a toolbox for small perturbations and adversarial attacks.

\subsection{Related works}



Fooling learning algorithms, and specifically deep neural networks, has been extensively studied in literature. Small agnostic modification of the inputs~\cite{hendrycks2019benchmarking} can already highly decrease accuracy of these algorithms. Adversarial perturbations~\cite{goodfellow2014explaining} make it even worse for the algorithms. At test time, it is thus possible to design a specific marginal perturbation such as a targeted network eventually predicts different outputs on original and disturbed input. Adversarial examples threats have been recently studied in remote sensing~\cite{chen2019adversarial}. However, adversarial perturbation is not a threat in many non safety-critical remote sensing applications as very large physical changes are required to produce remote sensing adversarial examples. Yet, one could still wonder if geographic generalization or over-fitting resistance could be helped by adversarial defenses.

There are currently two main adversarial defenses. First, adversarial retraining~\cite{shaham2018understanding} consists in finetuning the network using worse possible adversarial example at each iteration. Similar to a data augmentation process, this can lead to provable defense \cite{zhang2019theoretically} and corresponds somehow to learn the network in a support vector margin framework.  Second, adversarial regularization~\cite{bartlett2017spectrally} relies on Lipschitz constants to lower the neural networks adversarial sensibility. 

\section{Methodology}
\begin{table*}[!h]
\begin{minipage}{.34\linewidth}
\begin{center}
\begin{tabular}{c|ccc}
\toprule
 \multirow{2}{*}{\diagbox[]{Test}{Train}}& AIRS  & AIRS  & AIRS\\
 & \textit{naive} & \textit{add} & \textit{retr} \\
\midrule
AIRS  & 74 & \textbf{85} & 72\\
AIRS \textit{adv} & 63 & \textbf{77} & 68\\ \hline
ISPRS  & 58 & \textbf{80} & 62 \\
ISPRS \textit{adv} & 44 & \textbf{69} & 54 \\
\bottomrule
\end{tabular}
\caption{\label{segmentation} Root segmentation IoU depending on train/test data.
Test data can be with (\textit{adv}) or without adversarial attacks. Training can be \textit{naive}, with adversarial defense (\textit{retr}) or with extra data \textit{(add)}.}
\end{center}
\end{minipage}\hfill
\begin{minipage}{.62\linewidth}

\begin{center}

\begin{tabular}{c|ccc | c  c c }
\toprule
 \multirow{2}{*}{\diagbox[]{Model}{Data}} & VEDAI  & DFC  & ISPRS & VEDAI & DFC & ISPRS\\
 & \textit{32$\times$32} & \textit{48$\times$48} & \textit{64$\times$64} & \textit{adv} & \textit{adv} & \textit{adv}

\\ \midrule
VGG & 53 & 66 & 78 & 30 & 51 & 66\\
VGG \textit{reg} & 53 & 65 & 80 & \textbf{36} & \textbf{57} & \textbf{74}\\
VGG \textit{add} & \textbf{62} & \textbf{71} & \textbf{85} &31 & 49 & 64\\\hline
RESNET & 39 & 60 & 80 & 22 & 24 & 50\\
RESNET \textit{reg} & 43 & 49 & 81 & 20 & \textbf{29} & \textbf{69}\\
RESNET \textit{add} & \textbf{57} & \textbf{65} & \textbf{82} & 21 & 26 & 52\\
\bottomrule
\end{tabular}

\caption{\label{detection} G-score performances of a naive detector, an adversarial defended (\textit{reg}) one and one with extra data (\textit{add}) on clean and adversarial (\textit{adv}) datasets. \textit{Target size in different dataset is indicated in italic for completeness.}}
\end{center}
\end{minipage}\hfill
\end{table*}
\label{sec:methodo}
\subsection{Purpose}


Given the major influence of adversarial attacks, an idea could be to use adversarial defense as a way to increase general robustness in addition to the adversarial one.
Our overall purpose is thus to evaluate under adversarial defense the correlation between adversarial robustness and robustness to geographic transfer and over-fitting. To better apprehend the relevance of this approach, we increase the amount of training data with additional data for comparison. Hence, we also analyze the impact of a larger training database on adversarial attacks.

\subsection{Attack and defense}
\subsubsection{Adversarial retraining}
We consider the Fast Sign Gradient Method~(FSGM) adversarial attack~\cite{papernot2016limitations} and base our adversarial retraining scheme on this algorithm. FSGM is implemented by maximizing the loss over the image. The loss is averaged over the pixels in segmentation and over the targets in detection.
Formally, given an image $x$, its associated ground-truth $y$, a loss $\mathcal{L}$ and a trained neural network $f_\theta$ parameterized by $\theta$, FSGM considers the solution $\varepsilon = \lambda sign(\nabla_x \mathcal{L}(f_\theta(x),y))$ to the following problem :
\begin{equation}
    \underset{\varepsilon \ / \ ||\varepsilon||_1\leq \delta}{\max} \ \mathcal{L}(f_\theta(x+\varepsilon),y)
\end{equation}
$\delta$ is fixed such that only a dense modification of each pixel by 2 over 255 is possible.
Consistently in all of our experiments, FSGM attack is more effective than agnostic attacks like blur or pepper and salt noise.
\subsubsection{Adversarial regularization}
A function $f$ from $\mathbb{R}^I$ to $\mathbb{R}^J$ is said K-Lipschitz for norms $\|.\|_I$ and $\|.\|_J$ if there exists a constant $K\in\mathbb{R}$ such that, for all $x,y\in\mathbb{R}^I$, $\|f(x)-f(y)\|_J\leq K\|x-y\|_I$. As piece-wise
infinitely derivable functions, ReLu-based deep network are K-Lipschitz functions and could thus be naively expected to be smooth. However, both $K$ and $I$ can be very large: indeed, $K$ is estimated to by higher than 50000 for first Alexnet layers in $L_2$ norm~\cite{Szegedy13}. As sufficiently small Lipschitz coefficient would lead to less adversarial sensibility~\cite{huster2018limitations}, we implement an idea close to~\cite{bartlett2017spectrally} to this aim. Mainly, we add the following regularization term to the classical cross entropy: $|| \nabla_x ||f(x) - \hat{f}||_2^2 ||_1$ where $f$ is the last feature map of the encoder part of the network and $\hat{f}$ the average of $f(x)$ on the batch.

Code is available for implementation details.


\subsection{Experimental set-up}
\subsubsection{Data}
Experiments have been conducted on multiple public remote sensing datasets for both object detection and segmentation tasks: AIRS~\cite{chen2018aerial}, INRIA~\cite{huang2018large}, ISPRS (Potsdam and Vaihingen)~\cite{potsdam}, IEEE GRSS DFC~\cite{lagrange2015benchmarking} and VEDAI~\cite{razakarivony2016vehicle} datasets both for segmentation and detection.
All data are gray-scaled to be invariant to color channels. This is especially relevant for Vaihingen which is an IR-R-G dataset, all the others being RGB but with different distributions.

These datasets provide a wide range of use cases: AIRS and INRIA address building detection while VEDAI is only for vehicle detection. On the other hand, DFC and ISPRS are smaller but more versatile since they consist in multi class semantic segmentation.
Finally, they also provide a large range of resolutions, going from 5 cm/pixel for Potsdam to 30 cm/pixel for INRIA. The impact of resolution is also considered in the experiments.

\subsubsection{Backbone}
For detection, detectors are based on the SSD framework~\cite{liu2016ssd} with either VGG or ResNet backbone.
For segmentation, experiments rely on a UNet~\cite{ronneberger2015u} architecture but consistent results have been observed with Deeplabv3+.

\subsubsection{Metric}
Intersection over Union (IoU) is used to evaluate the performances in all the segmentation experiments. 

For detection, the classic way to match predicted bounding boxes and ground-truth bounding boxes is an Intersection over Union above some fixed threshold.
However, since scale is not an issue when resolution is known, we follow~\cite{Audebert_2017} and match predictions and ground-truth when centers are closer than 1.5 meter.
Then, we rely on F and G-scores to evaluate detection performances: F-score is the harmonic mean of precision and recall while G-score is the geometric one. Precision is the number of true alarms divided by the number of alarms and recall is the number of true alarms divided by the number of true objects. 

\section{Results}
\label{sec:results}
\subsection{Segmentation results}

Segmentation results are presented in Table~\ref{segmentation}. In addition to these results, we observed an IoU of 80\% on ISPRS when the network is trained on the ISPRS train set. 
We use adversarial retraining as the adversarial defense for this experiment since we observed better results than adversarial regularization. The additional training set consists of INRIA and DFC datasets. 

Consistently with adversarial works, the performances of the naive classifier decrease importantly under adversarial attack: they respectively drop by 11\% and 14\% on AIRS and ISPRS. Regarding these attacks, adversarial defense allows to make the network very resilient. Indeed, IoU only decreases by 6\% on average under adversarial attacks for the defended network against 12.5\% for the naive one. 

However, this adversarial robustness does not provide any geographic robustness. Indeed, IoU highly decreases when the model trained on the train set of AIRS is applied on ISPRS Vaihingen. This phenomenon is independent of the use of adversarial examples: IoU goes from 74\% to 58\% for the naive network and from 72\% to 62\% for the defended one. 

Inversely, adding additional data from two other datasets to the training set decreases overfitting and transfer capability. Indeed, performance increases both on the test set AIRS and on ISPRS, going respectively from 74\% \& 58\% to 85\% \& 80\%. It thus almost catches up on ISPRS with the control network trained on ISPRS.  Nonetheless, it does not protect against adversarial attacks: IoU still drops under attack by about 10\% both on AIRS and ISPRS. 

Hence, since additional data increases geographic robustness only while adversarial training increases adversarial robustness only, we conclude that adversarial and geographic robustness are uncorrelated in this experiment. 
\subsection{Detection results}
Detection results are reported in Table~\ref{detection}. 
Contrary to the previous segmentation experiment, pure transfer leads to extremely low performance in detection, so we consider here the impact of additional data and adversarial defense on classical test set performance: we study the impact on over-fitting. Thus, for a given dataset, we train a network classically, another one with adversarial defense and a last one with additional data from ISPRS Potsdam and DFC datasets and compare them on the associated test set. 
Finally, we base the adversarial defense here on adversarial regularization instead of retraining for computational reasons.

Like in the previous experiment, native models are very sensitive to FGSM. Indeed, their performance under adversarial attacks are extremely low compared to their performances on clean data: we observe G-score drops up to 30\%. Adding adversarial defense again importantly moderates this drop. Regularized algorithms eventually outperform their undefended version on adversarial data by 5\% of G-score. However, adversarial defense has no effect on clean data. Inversely, the additional training data clearly improves the performance on clean data but does not provide any defense against FGSM attack.

To summarize, Table~\ref{detection} clearly shows that adding data outperforms the baseline and the defended baseline on clean data whilst the defended baseline outperforms the other ones under adversarial attack. This shows a lack of correlations between adversarial robustness and robustness to over-fitting.

\section{Conclusion}
\label{sec:conclu}
This article aims to evaluate whether adversarial defense would be relevant data augmentation in a context where adversarial threats would be non-existent.
Hence, it focuses on the search of a correlation between adversarial, geographic and over-fitting robustness. Specifically, we investigate this concern for both object detection and segmentation, with different backbones and adversarial defense. 

Consistently in our experiments, adversarial defense does not improve both geographic and over-fitting robustness even though it strengthens the network against adversarial attacks. Therefore, using adversarial data as data augmentation does not seem relevant in an adversarial attack free context and under-performs compared to other methods such as the use of additional data. Inversely, adversarial defense greatly improves performance under adversarial attack, contrary to the use of extra datasets. 
To conclude, it seems that adversarial defense framework has little relevance in remote sensing applications without adversarial examples threats.




\bibliographystyle{IEEEbib}
\bibliography{bibli}

\end{document}